\documentclass{amia}
\usepackage{graphicx}
\usepackage[labelfont=bf]{caption}
\usepackage[superscript,nomove]{cite}
\usepackage{color}
\usepackage{amsmath}

\usepackage{algorithm}
\usepackage{algorithmic}

\usepackage{array}

\begin{document}

\title{A Novel Independent RNN Approach to Classification of Seizures against Non-seizures}

\author{Xinghua Yao, Ph.D$^{1}$, Qiang Cheng, Ph.D$^{1}$, Guo-Qiang Zhang, Ph.D$^{2}$}

\institutes{
    $^1$Institute of Biomedical Informatics, University of Kentucky, Lexington, Kentucky, USA; $^2$The University of Texas Health Science Center at Houston, Houston, Texas, USA \\
}

\maketitle

\noindent{\bf Abstract}

\textit{In current clinical practices, electroencephalograms (EEG) are reviewed and analyzed by trained neurologists to provide supports for therapeutic decisions. Manual reviews can be laborious and error prone. Automatic and accurate seizure/non-seizure classification methods are desirable. A critical challenge is that seizure morphologies exhibit considerable variabilities. In order to capture essential seizure features, this paper leverages an emerging deep learning model, the independently recurrent neural network (IndRNN), to construct a new approach for the seizure/non-seizure classification. This new approach gradually expands the time scales, thereby extracting temporal and spatial features from the local time duration to the entire record. Evaluations are conducted with cross-validation experiments across subjects over the noisy data of CHB-MIT. Experimental results demonstrate that the proposed approach outperforms the current state-of-the-art methods. In addition, we explore how the segment length affects the classification performance. Thirteen different segment lengths are assessed, showing that the classification performance varies over the segment lengths, and the maximal fluctuating margin is more than 4\%. Thus, the segment length is an important factor influencing the classification performance.
}

\section{Introduction}
More than 50 million people in the world suffer from epilepsy\cite{Megiddo}. Epilepsy is a central nervous system disorder, in which brain activity becomes abnormal, leading to sensations and sometimes loss of awareness. It can be life-threatening. Patients with epilepsy bear a high burden from disease in their daily lives, for example, having stringent limitations of acquiring and using a driving license\cite{Elger}. An important technique to diagnose epilepsy is to use electroencephalography (EEG). EEG records the electrical activities of the brain, and may reveal patterns of normal or abnormal brain electrical activities. In current clinical practices, EEG signals are collected from the brains by making use of either non-intrusive or implanted devices. The collected EEG signals are then reviewed and analyzed by trained neurologists to identify characteristic patterns of the disease, such as seizures and pre-ictal spikes. Disease information, like seizure frequency, seizure type, etc., are to provide supports for therapeutic decisions. However, this manual handling is tedious and error prone, and takes several hours for a trained professional to analyze one-day recordings from one patient\cite{Gotman1982,Thodoroff,Furbass,Zandi,Shoeb2010,Shoeb2009}. These limitations have motivated researchers to develop automated approaches to recognize seizures. In this paper, we focus on developing an automatic approach to classify seizure segments from the off-line EEG data for assisting physicians in making diagnosis.

Inter-patient and intra-patient seizure-morphology variation is a source of technical challenge in machine learning approaches to seizure/non-seizure classification of EEG signals acquired from epilepsy patients. Different machine learning methods and computational technologies have been applied to address this challenge. There are extensive studies for constructing patient-specific detectors capable of detecting seizure onsets \cite{Zandi,Shoeb2010,Amin,Hunyadi,Esbroeck,Vidyaratne}. In these studies, the problem of seizure detection is often converted into the seizure/non-seizure classification problem but more of a real-time flavor. Using traditional machine learning methods, hand-crafted features are usually needed to capture characteristics of seizure manifestations in EEG. More recent studies focus on designing deep learning approaches for seizure detection \cite{Thodoroff,Vidyaratne,Truong2018,Golmohammadi,Acharya}. There are components shared by most of these studies. For example, signal processing techniques are used to filer data; modules need to be pre-trained; multiple channels are utilized to extract spatial features; temporal features are extracted by sliding windows, and so on. Most of the deep learning-based approaches for seizure detection are developed based on classical neural network models, like convolutional neural network (CNN), recurrent neural network (RNN), long short-term memory (LSTM) and gated recurrent unit (GRU). While widely used, these standard neural network models have limitations for handling EEG data. CNN is good at processing two or more dimensional data, while EEG data are one dimensional and thus are not directly suited to CNN; RNN usually suffers from the gradient vanishing or exploding problem; though LSTM and GRU improved upon RNN, the training of a deep LSTM or GRU based network is in general difficult because of gradient decay over layers\cite{Shuai_Li}. An emerging variant of RNN, independently recurrent neural network (IndRNN), addresses the above limitations. By taking the Hadamard product over the recurrent inputs  \cite{Shuai_Li}, it overcomes the gradient vanishing or exploding problems, and supports computations over multiple layers efficiently. Additionally, it is able to process longer sequence data than LSTM. To exploit such advantages, this paper leverages IndRNN to design a new approach for the seizure/non-seizure classification.

EEG signals are highly dynamic and non-linear\cite{Andrzejak}. From different brain areas they have different morphologies\cite{Shoeb2010,Andrzejak}. And also seizure patterns in EEG data may manifest different temporal and spatial characteristics that have long-range correlations. Based on these observations, we propose an approach which extracts temporal features ranging from local time durations to the entire record. Spatial features are extracted from EEG signals from different brain areas. At the smallest time scale, IndRNN is utilized to extract temporal features at each time step. With several loops of IndRNN and max-pooling, the scale of time duration increases gradually, and the temporal features cover a longer time duration. Then, the overall features are extracted by an average pooling operation. The extracted features are passed into two fully connected layers for further integration and for final classification. In order to reduce the scale differences and speed up training, a batch normalization layer is inserted after each IndRNN layer. We perform extensive cross-validation experiments across subjects to evaluate the proposed approach. In experiments, we obtain promising results that are comparable or superior to those of the current state-of-the-art approaches. In addition, we explore how the segment length affects the performance of seizure/non-seizure classification. Our experimental results with different segment lengths show that the performance of seizure recognition fluctuates over the segment lengths, and the maximal fluctuating margin is more than 4\%.

The main contributions of our paper include the following: (1) An emerging deep learning model, IndRNN, is applied to seizure/non-seizure classification for the first time, and multi-scale temporal features are extracted with a deep architecture; (2) The relationship between the segment length and the performance of seizure/non-seizure classification is investigated; (3) Extensive cross-validation experimental results on the noisy EEG data of CHB-MIT demonstrate that seizures can be recognized more accurately, and the inter-patient seizure variabilities can be better overcome than current state-of-the-art deep learning approaches.

\section{Related Work}
Seizure/non-seizure classification distinguishes seizure segments from non-seizure segments, which can be used to recognize whether a data segment contains seizure or not. For this task, extensive studies have been performed. Because seizure detection, which is often of a real-time flavor, is often treated as the seizure/non-seizure classification problem, many machine learning methods have been developed \cite{Shoeb2010,Amin,Hunyadi,Esbroeck,Vidyaratne,Fergus,Nicalaou,Kharbouch,Bolagh}. Recently, deep learning techniques have been applied to the seizure detection problem\cite{Thodoroff,Truong2018,Golmohammadi,Acharya,Hussein}. The evaluations of these methods are conducted with patient-specific or across-patients experiments. The classification across patients is more challenging because it needs to overcome the inter-patients variabilities.

Shoeb and Guttag propose a method to construct a patient-specific detector for seizure detection by using the support vector machine (SVM)\cite{Shoeb2010}. The method leverages filters to extract spectral features over each channel, and then stacks feature vectors to catch time-evolution information. A  sensitivity of 96\% is achieved with the mean latency of 4.6s, and the median false positive rate is 2 false detections per 24 hours. The performance results are often used as a benchmark for patient-specific seizure detection on the data set CHB-MIT.

Zandi et al. propose a wavelet-based algorithm for real-time detection of epileptic seizures using scalp EEG\cite{Zandi}. In this algorithm, the EEG from each channel is decomposed by wavelet packet transform, and a patient-specific measure is deployed by using wavelet coefficients to separate the seizure and non-seizure states. Utilizing the measure, a combined seizure index is derived for each epoch of every EEG channel. Through inspecting the combined seizure index, proper channel alarms are generated.

Fergus et al. present a method for the seizure/non-seizure classification based on traditional machine learning techniques, and obtain
88\% in sensitivity and 88\% in specificity over CHB-MIT\cite{Fergus}. The method consists of data filtering, feature extraction, feature selection and training classifiers. It is evaluated over data segments which are produced by the authors' proposed segmentation method. A seizure segment is only truncated from the beginning of one seizure and the truncated length is 60 seconds. Non-seizure segment is truncated from non-seizure records. On the average, each seizure segment contains 40s ictal data.

Vidyaratne et al. propose a deep recurrent architecture by combining cellular neural network and bidirectional RNN\cite{Vidyaratne}. The bidirectional RNN is deployed into each cell of the cellular neural network to extract temporal features in the forward and the backward directions. Each cell interacts with its neighboring cells to extract local spatial-temporal features. The sensitivities are 100\% in patient-specific experiments over five patients from CHB-MIT.

Thodoroff et al. design a recurrent convolutional neural network to capture spectral, spatial and temporal patterns of
seizures\cite{Thodoroff}. EEG signals are firstly transformed into images. Created images are fed into CNN. Output vectors of the CNN are organized to be sequences in chronological order. The sequences are passed into the bidirectional RNN to make classification. Both patient-specific experiments and cross-patient experiments are conducted. In the cross-patient testing, the sensitivity is 85\% on average and the false positive rate is 0.8/hours. Transfer learning technique is utilized to overcome the problem of small amount of data in the patient-specific experiments.

Golmohammadi et al. explore two kinds of neural networks over TUH EEG Corpus\cite{Golmohammadi}. Their experiment results show that convolutional LSTM network outperforms convolutional GRU network. Different initialization and regularization methods are considered. LSTM and GRU are limited when using multiple layers.

Hussein et al. design a deep neural network for seizure/non-seizure classification by using LSTM\cite{Hussein}. It extracts temporal features by using LSTM. Acharya et al. present a 13-layers deep neural network for seizure/non-seizure classification by using CNN\cite{Acharya}. The two approaches are evaluated over the same EEG data set provided by University of Bonn\cite{Andrzejak}. The LSTM approach achieves performances of 100\%. The CNN approach obtains the sensitivity of 95\% and the specificity of 90\%. Each record in the Bonn EEG data set contains only one channel and has no artifacts.

These developed seizure-detection methods based on traditional machine learning techniques can work well with small amount of samples. They often need
crafted features and manually selecting features. In the currently developed seizure/non-seizure classification methods based on deep learning techniques, most methods are based on classical neural network models, such as CNN, RNN, and LSTM. These models have their limitations for processing EEG signal data, such as suffering from gradient vanishing or exploding problem.

\section{Method}

\subsection{Model Design}
EEG signals are dynamic\cite{Andrzejak}, and seizure morphologies varies with brain areas\cite{Shoeb2010,Andrzejak}. Capturing temporal-spatial features in EEG signals is critical to seizure/non-seizure classification. For extracting spatial features, EEG signals over multiple channels are taken as inputs. For capturing overall temporal features, we leverage IndRNN in different time scales. Predominant features in local time scales are computed with max-pooling operation. Sequence produced by the combination of IndRNN and max-pooling operation forms a new time scale hierarchy. Stacking such combinations provides temporal features. Local features in the last hierarchy are averaged as overall features. Finally, the overall features are passed into fully connected layers to integrate features and make final classification. Batch normalization is utilized after IndRNN to facilitate training.

\subsection{Model Architecture}
\begin{figure}[ht]
\centering
\includegraphics[scale=0.48]{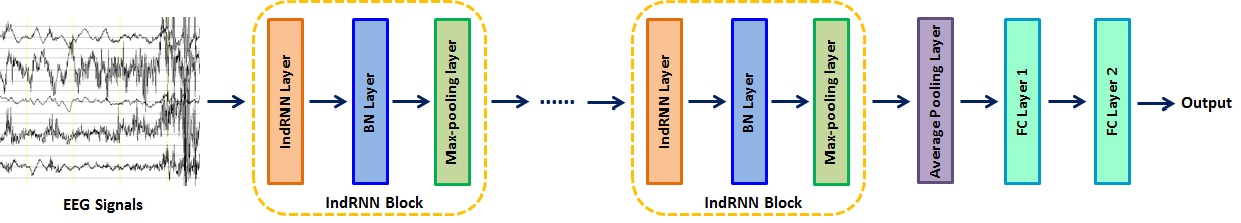}
\caption{Architecture of IndRNN approach}
\label{architecture-IndRNN-approach}
\end{figure}

Our model architecture consists of IndRNN blocks, one average pooling layer and two fully connected layers. Each IndRNN block comprises of an IndRNN layer, a bach normalization layer, and a max pooling layer. The architecture is depicted in Figure \ref{architecture-IndRNN-approach}.

\begin{itemize}

\item \textit{\textbf{IndRNN layer:}} IndRNN layer processes input sequences in forward order, and extracts time-dependent features. It executes computation as follows:
\begin{align}
    \label{eq-hidden-layer-IndRNN} \mathbf{h}_{t} & =  \sigma_{\mathbf{h}}(\mathbf{W}_{\mathbf{h}}*\mathbf{x}_{t} + \mathbf{u}_{\mathbf{h}}\odot\mathbf{h}_{t-1} + \mathbf{b}_{\mathbf{h}}) \\
    \label{eq-output-IndRNN} \mathbf{y}_{t} & = \sigma_{\mathbf{y}}(\mathbf{W}_{\mathbf{y}} * \mathbf{h}_{t} + \mathbf{b}_{\mathbf{y}})
\end{align}
Here, $\mathbf{x}_{t}$, $\mathbf{h}_{t}$, $\mathbf{y}_{t}$ are input vector, hidden layer vector, and output vector at the time $t$, respectively. $\mathbf{W}_{\mathbf{h}}$ is an input weight matrix, $\mathbf{u}_{\mathbf{h}}$ is recurrent weight vector, and $\mathbf{W}_{\mathbf{y}}$ is an output weight matrix. $\mathbf{b}_{\mathbf{h}}$ and $\mathbf{b}_{\mathbf{y}}$ are bias weights. $\odot$ represents Hadamard product. $\sigma_{\mathbf{h}}$ and $\sigma_{\mathbf{y}}$ are activation functions such as ReLU.

\item \textit{\textbf{BN layer:}} It is inserted after each IndRNN layer. It is used to speed up training and to reduce overfitting\cite{Sergey_Ioffe}.

\item \textit{\textbf{Max-pooling layer:}} In each IndRNN block, a max-pooling layer is applied to the batch normalized results. It extracts predominant features from the normalized sequences at a specific temporal scale.

\item \textit{\textbf{Average pooling layer:}} Following IndRNN blocks, an average pooling layer is inserted to extract overall features across time scales for the final classification.

\item \textit{\textbf{FC layers:}} Two fully connected layers are designed. The first aims to integrate features from the outputs of the average pooling layer over channels and make further extractions. The second is to perform final classification of seizure/non-seizure.

\end{itemize}


\section{Evaluation}

\subsection{Data Set}
The data set of CHB-MIT\cite{Shoeb2009} contains 686 EEG recordings from 23 subjects of different ages ranging from 1.5 years to 22 years. The recordings include 198 seizures. The sampling frequency is 256 Hz. Most recordings are one hour long, and some are two hours long or four hours long. The EEG recordings are grouped into 24 cases. In each case, the data recordings are from a single subject. Case chb21 was obtained 1.5 years after Case chb01 from the same subject. Each data file contains data over 23 or more channels. In several data files, there are missing values; thus, we only consider those channels without missing values. Three data files, including chb12\_27.edf, chb12\_28.edf and chb12\_29.edf, have different channel montages from other data files. In our experiments, we remove these three data files.

\subsection{Data Segmentation}
In order to extract effective seizure features, 17 common channels are chosen. According to a data segment length of 23 seconds, each data record in each case is split into data segments from the beginning to the end without overlapping. If the duration of a data record is not divided by the segment length and there is a seizure happening in the remaining part, we will ensure that the last segment will have the same length but will overlap with its prior segment. If the remaindering part contains no seizure, then it is dropped. Using annotation files for this data set, we determine whether a data segment contains a seizure or not. In our experiments, if a segment contains any seizure data, it is considered as a seizure segment; otherwise, it is a non-seizure segment.

Using the above segmentation method, 665 seizure segments are obtained. In these seizure segments, the lengths of seizures vary from 1s to 23s, with the average length being 16.9s. Among all seizure segments, segments containing seizure signals of less than 7s comprise 14.7\%, those containing more than 17s comprise 59.8\%, and those containing more than 10s comprise 76.1\%. All the seizure data segments are taken as a part of our experiment data. We randomly choose 665 non-seizure segments in each experiment. The 1330 seizure/non-seizure segments are randomly split into training set, validation set and testing set with a ratio of 70:15:15 in each experiment, and we adopt the repeated random sub-sampling validation as a strategy for cross validation.




\subsection{Cross-Validation Results for the Proposed Approach}

Based on the architecture in Figure \ref{architecture-IndRNN-approach}, we build a model by stacking 15 IndRNN layers to classify seizure/non-seizure segments from CHB-MIT. In the model, the main parameters are set as follows: The number of hidden states in the first five IndRNN layers is 128, that in the second five IndRNN layers is 200, that in the third five IndRNN layers is 250, each layer for max pooling has a window size of 2 and stride of 2, the number of hidden states for the first fully connected layer is 100, the optimizer is Adam, the learning rate is 0.0004. We train the model using a batch size of 30 for 100 epochs in each experiment. Overall, ten cross-validation experiments are conducted. The obtained results of these ten experiments are given in Table \ref{cross-validation-results-IndRNN}. It can be seen that the average (Ave.) in sensitivity is 87.3\%, in specificity is 86.7\%, in precision in 87.08\%, and in F1 score is 87.07\%.
\begin{table}[h]
\centering
\caption{Cross-validation results using the proposed approach}
\label{cross-validation-results-IndRNN}
\begin{tabular}{|l||l|l|l|l|l|}
\hline
\textbf{Item} & \textbf{Sensitivity} & \textbf{Specificity} & \textbf{F1 Score} & \textbf{Precision} & \textbf{Accuracy} \\
\hline
1 & 0.9100 & 0.8300 & 0.8750 & 0.8426 & 0.8700 \\
\hline
2 & 0.8900 & 0.9000 & 0.8945 & 0.8990 & 0.8950  \\
\hline
3 & 0.9300 & 0.8600 & 0.8986 & 0.8692 & 0.8950 \\
\hline
4 & 0.7900 & 0.8500 & 0.8144 & 0.8404 & 0.8200 \\
\hline
5 & 0.8400 & 0.8900 & 0.8615 & 0.8842 & 0.8650 \\
\hline
6 & 0.8500 & 0.8600 & 0.8543 & 0.8586 & 0.8550 \\
\hline
7 & 0.8700 & 0.8500 & 0.8614 & 0.8529 & 0.8600\\
\hline
8 & 0.8700 & 0.9500 & 0.9062 & 0.9457 & 0.9100 \\
\hline
9 & 0.9000 & 0.7300 & 0.8295 & 0.7692 & 0.8150 \\
\hline
10 & 0.8800 & 0.9500 & 0.9119 & 0.9462 & 0.9150 \\
\hline
\hline
Ave. & 0.8730 & 0.8670 & 0.8707 & 0.8708 & 0.8700 \\
\hline
Std. & 0.0377 & 0.0602 & 0.0310 & 0.0498 & 0.0328 \\
\hline

\end{tabular}
\end{table}

\subsection{Comparison with the LSTM and CNN Approaches}

As a main module LSTM has been used to detect seizures\cite{Hussein}. The LSTM approach is evaluated through cross-validation experiments over the EEG data set from University of Bonn\cite{Andrzejak}, showing state-of-the-art performance. A CNN-based approach has also been proposed for seizure/non-seizure classification\cite{Acharya}, which also demonstrates state-of-the-art performance over the Bonn data set. Because the Bonn data set is heavily processed and contains no artifacts, and its size is small, we compare the proposed approach with the LSTM approach and the CNN approach over the noisy dataset CHB-MIT.

For the LSTM approach and the CNN approach, we implement them according to their descriptions in the related literatures. And the implementations are tested. Our obtained testing results reach the reported performances. Based on the two implementations, cross-validation experiments are conducted for the LSTM approach and the CNN approach separately. The LSTM approach consists of one LSTM layer, one time-distributed computing layer, one average pooling layer and one fully connected layer.
In the experiments using the LSTM approach, our parameter setting is as follows: The number of hidden states is 120 in the LSTM layer, that in the time-distributed computing layer is 60, the optimizer is RMSprop, the learning rate is 0.0007, the batch size is 30, and epochs is 30.
For the CNN approach, it contains five convolutional layers, five max pooling layers, and three fully connected layers. The parameters are set as follows: the number of hidden states in the first two convolutional layers is 100, that in each of the second two convolutional layers is 200, that in the fifth convolutional layer is 260, that in the first fully connected layer is 100, that in the second fully connected layer is 50, the parameter alpha is 0.01 in the LeakyReLU activation function, the optimizer is Adam, the learning rate is 0.001, the batch size is 30, and the number of epochs is 50. Using the LSTM approach, ten cross-validation results are obtained and shown in Table \ref{cross-validation-results-LSTM}. The obtained average sensitivity is 84.4\%, the average specificity is 84.3\%, and the average precision is 84.7\%. For the CNN approach, ten cross-validation experiments are conducted, and the results are given in Table \ref{cross-validation-results-CNN}. The achieved average sensitivity, the average specificity and the average precision are 84.8\%, 81.0\% and 82.56\%, respectively.
\begin{table}[h]
\centering
\caption{Cross-validation results using the LSTM approach}
\label{cross-validation-results-LSTM}
\begin{tabular}{|l||l|l|l|l|l|}
\hline
\textbf{Item} & \textbf{Sensitivity} & \textbf{Specificity} & \textbf{F1 Score} & \textbf{Precision} & \textbf{Accuracy} \\
\hline
1 & 0.8500 & 0.8800 & 0.8629 & 0.8763 & 0.8650 \\
\hline
2 & 0.7700 & 0.8500 & 0.8021 & 0.8370 & 0.8100 \\
\hline
3 & 0.7900 & 0.8700 & 0.8229 & 0.8587 & 0.8300 \\
\hline
4 & 0.7100 & 0.9300 & 0.7978 & 0.9103 & 0.8200 \\
\hline
5 & 0.8200 & 0.8900 & 0.8497 & 0.8817 & 0.8550 \\
\hline
6 & 0.9100 & 0.7900 & 0.8585 & 0.8125 & 0.8500 \\
\hline
7 & 0.8600 & 0.8300 & 0.8473 & 0.8350 & 0.8450 \\
\hline
8 & 0.8600 & 0.8400 & 0.8515 & 0.8431 & 0.8500 \\
\hline
9 & 0.9400 & 0.7200 & 0.8468 & 0.7705 & 0.8300  \\
\hline
10 & 0.9300 & 0.8300 & 0.8857 & 0.8455 & 0.8800 \\
\hline
\hline
Ave. & 0.8440 & 0.8430 & 0.8425 & 0.8470 & 0.8435  \\
\hline
Std. & 0.0696 & 0.0550 & 0.0259 & 0.0368 & 0.0201  \\
\hline

\end{tabular}
\end{table}

\begin{table}[h]
\centering
\caption{Cross-validation results using the CNN approach}
\label{cross-validation-results-CNN}
\begin{tabular}{|l||l|l|l|l|l|}
\hline
\textbf{Item} & \textbf{Sensitivity} & \textbf{Specificity} & \textbf{F1 Score} & \textbf{Precision} & \textbf{Accuracy} \\
\hline
1 & 0.8400 & 0.8500 & 0.8442 & 0.8485 & 0.8450 \\
\hline
2 & 0.9200 & 0.7700 & 0.8558 & 0.8000 & 0.8450 \\
\hline
3 & 0.8000 & 0.8400 & 0.8163 & 0.8333 & 0.8200  \\
\hline
4 & 0.9000 & 0.6900 & 0.8145 & 0.7438 & 0.7950  \\
\hline
5 & 0.9200 & 0.8000 & 0.8679 & 0.8214 & 0.8600  \\
\hline
6 & 0.7900 & 0.8500 & 0.8144 & 0.8404 & 0.8200  \\
\hline
7 & 0.6300 & 0.9700 & 0.7590 & 0.9545 & 0.8000 \\
\hline
8 & 0.8500 & 0.8700 & 0.8586 & 0.8673 & 0.8600 \\
\hline
9 & 0.8700 & 0.7700 & 0.8286 & 0.7909 & 0.8200 \\
\hline
10 & 0.9600 & 0.6900 & 0.8458 & 0.7559 & 0.8250 \\
\hline
\hline
Ave. & 0.8480 & 0.8100 & 0.8305 & 0.8256 & 0.8290 \\
\hline
Std. & 0.0891 & 0.0809 & 0.0301 & 0.0571 & 0.0217 \\
\hline

\end{tabular}
\end{table}

Comparing cross-validation results in Tables \ref{cross-validation-results-IndRNN}-\ref{cross-validation-results-CNN}, we can conclude that the average performance, in either one of the metrics of sensitivity, specificity, F1 score, precision, and accuracy,  using the proposed approach is at least 2\% greater than that using the LSTM approach or the CNN approach. The above comparisons show that the proposed approach outperforms the LSTM approach and the CNN approach in seizure/non-seizure classification.

\subsection{Validation of IndRNN Layers}

The performance of a deep learning model is typically affected by the number of layers. In this section, we investigate the performance of the proposed approach with different numbers of IndRNN layers. Four cases with various layers, i.e. 6 IndRNN layers, 9 IndRNN layers, 12 IndRNN layers, and 15 IndRNN layers, are tested separately. For each case, ten cross-validation experiments are conducted based on the tuned optimal parameters over 23s data segments from CHB-MIT. The cross-validation results of the four cases are summarized in Table \ref{experiment-results-of-different-number-IndRNN-layers}.
\begin{table}[h]
\centering
\caption{Cross-validation results using different number of IndRNN layers}
\label{experiment-results-of-different-number-IndRNN-layers}
  \begin{tabular}{|l|l|l|l|l|l|}
  \hline
    \textbf{IndRNN layers} & \textbf{Sensitivity} & \textbf{Specificity} & \textbf{F1 Score} & \textbf{Precision} & \textbf{Accuracy} \\
    \hline
    6 layers & 0.8420$\pm$0.0387  &  0.8890$\pm$0.0416  &  0.8622$\pm$0.0282  &  0.8851$\pm$0.0393  &  0.8655$\pm$0.0278   \\
    \hline
    9 layers & 0.8440$\pm$0.0338  & 0.8770$\pm$0.0560  & 0.8584$\pm$0.0177 & 0.8768$\pm$0.0498 & 0.8605$\pm$0.0204   \\
    \hline
    12 layers & 0.8520$\pm$0.0424 & 0.8730$\pm$0.0377 & 0.8608$\pm$0.0175 & 0.8723$\pm$0.0314 & 0.8625$\pm$0.0157 \\
    \hline
    15 layers & 0.8730$\pm$0.0377 & 0.8670$\pm$0.0602 & 0.8707$\pm$0.0310 & 0.8708$\pm$0.0498 & 0.8700$\pm$0.0328  \\
    \hline

  \end{tabular}
\end{table}


The results in Table \ref{experiment-results-of-different-number-IndRNN-layers} indicate that the sensitivities in the four cases increase as the number of the IndRNN layers increases, but the specificities decrease. The four accuracies are similar. The structure with 15 IndRNN layers has the best sensitivity, and the difference between the sensitivity and the specificity is small. In order to detect more seizures, we select the approach with 15 IndRNN layers and use it to compare with the LSTM approach and the CNN approach for the seizure/non-seizure classification.

\subsection{Effects of Segment Lengths}

In the following, we explore relationship between the segment length and the performance of seizure/non-seizure classification.

\begin{figure}[ht]
\centering
\includegraphics[scale=0.5]{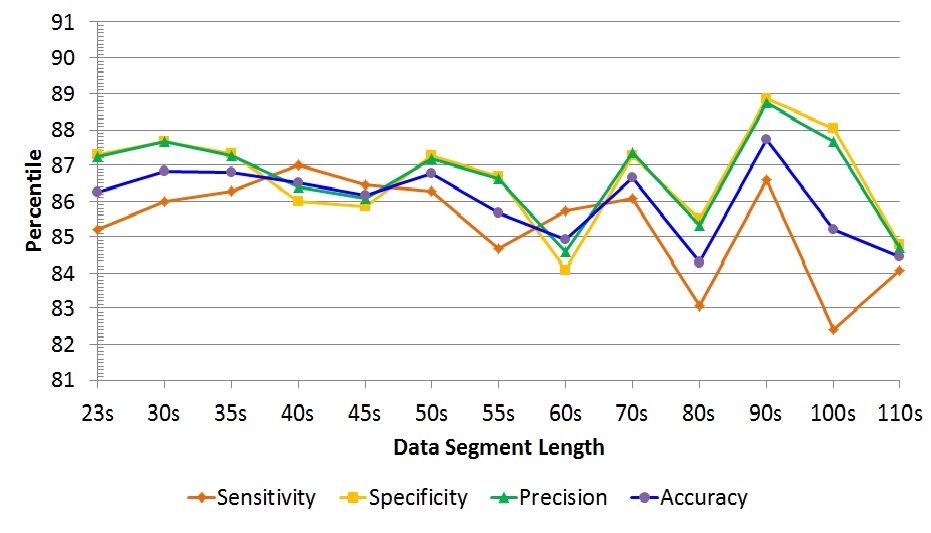}
\caption{Performance over data segments with different lengths}
\label{figure-results-over-segments-with-different-lengths}
\end{figure}

Generally, seizures last for less than two minutes. We select 13 temporal lengths less than 2 min and separately use each length to segment EEG signals in CHB-MIT. Besides the length of 23s, other 12 lengths are 30s, 35s, 40s, 45s, 50s, 55s, 60s, 70s, 80s, 90s, 100s, and 110s. The segmentation is similar to the case of 23s. For each length, ten cross-validation experiments are conducted based on a group of tuned optimal parameters by using the proposed approach containing 12 IndRNN layers. In the experiments, the tuned parameters mainly include the learning rate and the number of epochs. The same numbers of hidden states are used as in the case of 23s. The number of seizure segments for each segment length and the obtained cross-validation results are listed in Table \ref{expriments-over-segments-with-different-lengths} and visualized in Figure \ref{figure-results-over-segments-with-different-lengths}, where Len. stands for segment length, and Num. Sei. for the number of seizure segments.

\begin{table}[h]
\centering
\caption{Cross-validation results over segments with different lengths and number of produced seizure segments}
\label{expriments-over-segments-with-different-lengths}
  \begin{tabular}{|l|l|l|l|l|l|l|}
  \hline
    \textbf{Len.} & \textbf{Num. Sei.} & \textbf{Sensitivity} & \textbf{Specificity} & \textbf{F1 Score} & \textbf{Precision} & \textbf{Accuracy} \\
    \hline
   23s & 665 & 0.8520$\pm$0.0424 &  0.8730$\pm$0.0377 &  0.8608$\pm$0.0175 &  0.8723$\pm$0.0314 &  0.8625$\pm$0.0157  \\
   \hline
    30s &  543 & 0.8598$\pm$0.0427 &  0.8768$\pm$0.0398 &  0.8669$\pm$0.0186 &  0.8768$\pm$0.0329 &  0.8683$\pm$0.0175  \\
    \hline
    35s &  496 & 0.8627$\pm$0.0348 &  0.8733$\pm$0.0354 &  0.8673$\pm$0.0298 &  0.8725$\pm$0.0329 &  0.8680$\pm$0.0298  \\
    \hline
    40s &  463 & 0.8700$\pm$0.0370 &  0.8600$\pm$0.0541 &  0.8659$\pm$0.0283 &  0.8640$\pm$0.0463 &  0.8650$\pm$0.0293  \\
    \hline
    45s &  429 & 0.8646$\pm$0.0236 & 0.8585$\pm$0.0429 & 0.8622$\pm$0.0220 & 0.8609$\pm$0.0374 & 0.8615$\pm$0.0236  \\
    \hline
    50s &  411 & 0.8629$\pm$0.0300 &  0.8726$\pm$0.0233 &  0.8670$\pm$0.0204 &  0.8717$\pm$0.0208 &  0.8677$\pm$0.0198  \\
    \hline
    55s &  396 & 0.8467$\pm$0.0452 &  0.8667$\pm$0.0483 &  0.8550$\pm$0.0235 &  0.8664$\pm$0.0376 &  0.8567$\pm$0.0226  \\
    \hline
    60s &  358 & 0.8574$\pm$0.0414 &  0.8407$\pm$0.0505 &  0.8503$\pm$0.0221 &  0.8457$\pm$0.0371 &  0.8491$\pm$0.0227  \\
    \hline
    70s &  340 & 0.8608$\pm$0.0310 &  0.8726$\pm$0.0450 &  0.8660$\pm$0.0135 &  0.8735$\pm$0.0365 &  0.8666$\pm$0.0153  \\
    \hline
    80s &  325 & 0.8306$\pm$0.0429 &  0.8551$\pm$0.0423 &  0.8407$\pm$0.0261 &  0.8532$\pm$0.0363 &  0.8428$\pm$0.0255  \\
    \hline
    90s &  289 & 0.8659$\pm$0.0570 &  0.8886$\pm$0.0448 &  0.8754$\pm$0.0392 &  0.8874$\pm$0.0417 &  0.8773$\pm$0.0380  \\
    \hline
    100s &  305 & 0.8239$\pm$0.0579 &  0.8804$\pm$0.0569 &  0.8475$\pm$0.0383 &  0.8766$\pm$0.0528 &  0.8522$\pm$0.0368 \\
    \hline
    110s &  293 & 0.8409$\pm$0.0249 &  0.8477$\pm$0.0204 &  0.8437$\pm$0.0154 &  0.8470$\pm$0.0172 &  0.8443$\pm$0.0144 \\
    \hline

  \end{tabular}
\end{table}

Figure \ref{figure-results-over-segments-with-different-lengths} shows that the relation between the data segment lengths and the performances of seizure/non-seizure classification is not linear. The classification performance does not go up or go down as the segment length increases, and it fluctuates over the segment lengths. With six lengths, including 60s, 70s, 80s, 90s, 100s and 110s, the performances manifest wide fluctuating margin. With six lengths of 23s, 30s, 35s, 40s, 45s, and 50s, the differences of performances are relatively small. The best performance is obtained with the segment length of 90s. For the three metrics of Sensitivity, Specificity and Precision, their maximal gaps are all more than 4\%. For the F1 Score and Accuracy, the maximal differences are more than 3\%. It can be seen that the influence of the segment length can not be overlooked for the seizue/non-seizure classification. With different segment lengths, the distributions of time lengths of seizure in seizure segments are different. The fluctuations are likely to be from the differences between the distributions of seizure time lengths.

\section{Discussion}
 To automatically identify seizure segments from off-line EEG data for assisting neurologists in reviewing and analyzing, a deep learning approach is proposed to classify seizure against non-seizure. Compared to the LSTM approach and the CNN approach, our proposed approach has improvements typically of more than 2\%, and more than 4\% in specificity and precision. The improvements show the IndRNN approach is powerful to classify seizure/non-seizure on EEG data. The strength of the proposed approach is from the model IndRNN. IndRNN is able to handle many layers and longer sequences than LSTM and RNN\cite{Shuai_Li}. The IndRNN approach extracts features from the forward direction. When conducting experiments, we attempt to construct a bidirectional IndRNN approach to support the forward and the backward computing. Our experiments show that the bidirectional IndRNN approach is time-consuming in computation while the performance gain is marginal, and we thus adopt the unidirectional IndRNN approach.

When segmenting data, we attempt to ensure that the obtained data segments are close to a real-world scenario. A seizure segment could contain seizure data and non-seizure data. It is unrealistic in the real world that all the seizure segments only contain seizure data. As the lengths of seizure data in seizure segments vary significantly, the segment length of 23s is selected for evaluating the proposed approach against the LSTM and CNN approaches.

In the exploration of relations between segment lengths and performances of classifying seizure/non-seizure, we choose to use the IndRNN approach with 12 IndRNN layers. The choice is based on the following three considerations: (1) The structure with 12 IndRNN layers has relatively good sensitivity over the 23s segments, as shown in Table \ref{experiment-results-of-different-number-IndRNN-layers}; (2) For the structure with 15 IndRNN layers, more parameters need to be trained, and the number of seizure samples decreases as the segment length increases, which increases the risk of overfitting. (3) The results in Table \ref{expriments-over-segments-with-different-lengths} and Figure \ref{figure-results-over-segments-with-different-lengths} indicate that the influence of segment size is not small for the seizure/non-seizure classification.

\section{Limitations}
The proposed IndRNN approach processes EEG signals from different brain areas and extract spatial features. It does not distinguish signals over different channels in a strict way. The neural network in the proposed approach is deep and long. Its training needs much more samples than traditional machine learning methods, and its strength is limited in processing imbalanced data.

\section{Conclusion}
For seizure/non-seizure classification, a novel approach is proposed. The approach leverages an emerging neural network model, IndRNN, and achieves the state-of-art performance in cross-validation experiments across patients. The obtained performance in sensitivity, specificity, and precision, is better than LSTM and CNN approaches. The results demonstrate that our proposed approach is more resilient to the inter-patients variabilities and recognizes seizure segments more accurately. The proposed approach can accurately select most of seizure segments for assisting the neurologists in reviewing and analyzing. Additionally, we explore how the segment length affects the performance of seizure/non-seizure classification. Our cross-validation experiments over 13 segment lengths show that the classification performance fluctuates over the segment lengths, with the maximal fluctuating margin being more than 4\%. The segment length is thus an important factor influencing the seizure/non-seizure classification performance. As a future research line, we will investigate the problem of how to use the IndRNN approach for real-time seizure detection.

\makeatletter
\renewcommand{\@biblabel}[1]{\hfill #1.}
\makeatother

\bibliographystyle{unsrt}

\end{document}